\documentclass[10pt,twocolumn,letterpaper]{article}

\usepackage{arxiv}

\usepackage{times}
\usepackage{epsfig}
\usepackage{graphicx}
\usepackage{amsmath}
\usepackage{amssymb}
\usepackage{booktabs}
\usepackage{multirow, makecell}

\usepackage{algorithm}
\usepackage{algpseudocode}
\usepackage[pagebackref=true,breaklinks=true,letterpaper=true,colorlinks,bookmarks=false]{hyperref}
\usepackage{xcolor}
\usepackage{soul}
\usepackage{amsmath}
\usepackage{amssymb}
\usepackage{microtype}
\usepackage{wrapfig}

\def\figurePath{figures_arxiv/}
\def\myfigure#1#2{\begin{figure}[htb]\centering\includegraphics*[width = \linewidth]{\figurePath#1}\caption{#2}\label{fig:#1}\end{figure}}

\def\mycfigure#1#2{\begin{figure*}[t]\centering\includegraphics*[clip, width = \linewidth]{\figurePath#1}\caption{#2}\label{fig:#1}\end{figure*}}

\def\myfigurePlace#1#2#3{\begin{figure}[#3!]\centering\includegraphics[clip, width = \linewidth]{\figurePath#1}\caption{#2}\label{fig:#1}\end{figure}}

\renewcommand{\eg}{e.\,g., }
\renewcommand{\ie}{i.\,e., }
\renewcommand{\etal}{et~al.\ }
\newcommand{\etals}{et~al.'s}

\newcommand{\refSec}[1]{Sec.~\ref{sec:#1}}
\newcommand{\refFig}[1]{Fig.~\ref{fig:#1}}

\newcommand{\refTbl}[1]{Tbl.~\ref{tbl:#1}}
\newcommand{\refAlg}[1]{Alg.~\ref{alg:#1}}

\soulregister\ref7
\soulregister\cite7
\soulregister\refFig7
\soulregister\cite7
\soulregister\ref7
\soulregister\pageref7
\soulregister\shortcite7
\soulregister\eg0
\soulregister\ie0
\soulregister\etal0

\DeclareGraphicsExtensions{.png,.jpg,.pdf,.ai,.psd}
\newcommand{\mysection}[2]{\section{#1}\label{sec:#2}}
\newcommand{\mysubsection}[2]{\subsection{#1}\label{sec:#2}}

\usepackage{pifont}
\newcommand{\cmark}{\multicolumn{1}{c}{\checkmark}}%
\newcommand{\xmark}{\multicolumn{1}{c}{\scalebox{0.8}{\ding{53}}}}%

\usepackage{adjustbox}
\newcolumntype{R}[2]{%
    >{\adjustbox{angle=#1,lap=\width-(#2)}\bgroup}%
    l%
    <{\egroup}%
}

\newcommand*\rot[2]{\multicolumn{1}{R{#1}{#2}}}%

\newcommand{\name}{\textsc{PlatonicGAN}\xspace}
\renewcommand{\paragraph}[1]{
\vspace{.15cm}%
\noindent%
\textbf{#1\ }}

\MakeRobust{\Call}

\begin{document}

\title{Escaping Plato's Cave: 3D Shape From Adversarial Rendering}

\author{Philipp Henzler\\
{\tt\small p.henzler@cs.ucl.ac.uk}
\\
\and
Niloy J. Mitra\\
{\tt\small n.mitra@cs.ucl.ac.uk}\vspace{.3cm}\\
University College London\\
\and
Tobias Ritschel\\
{\tt\small t.ritschel@ucl.ac.uk}
\\
}

\maketitle

\begin{abstract}
We introduce \name to discover the 3D structure of an object class from an unstructured collection of 2D images, \ie where no relation between photos is known, except that they are showing instances of the same category.
The key idea is to train a deep neural network to generate 3D shapes which, when rendered to images, are indistinguishable from ground truth images (for a discriminator) under various camera poses.
Discriminating 2D images instead of 3D shapes allows tapping into unstructured 2D photo collections instead of relying on curated (\eg aligned, annotated, etc.) 3D data sets.

To establish constraints between 2D image observation and their 3D interpretation, we suggest a family of \emph{rendering layers} that are effectively differentiable.
This family includes visual hull, absorption-only (akin to x-ray), and emission-absorption.
We can successfully reconstruct 3D shapes from unstructured 2D images and extensively evaluate \name on a range of synthetic and real data sets achieving consistent improvements over baseline methods.
We further show that \name can be combined with 3D supervision to improve on and in some cases even surpass the quality of 3D-supervised methods.
\end{abstract}

\mysection{Introduction}{Introduction}
A key limitation to current generative models~\cite{wu20153d,wu2016learning,girdhar2016learning,qi2016volumetric,wang2017shape,wang2018deep} is the availability of suitable training data (\eg 3D volumes, feature point annotations, template meshes, deformation prior, structured image sets, etc.) for supervision.

While methods exist to learn the 3D structure of classes of objects, they typically require 3D data as input.
Regrettably, such 3D data is difficult to acquire, in particular for the ``long tail'' of exotic classes: ShapeNet might have \texttt{chair}, but it does not have \texttt{chanterelle}.

Addressing this problem, we suggest a method to learn 3D structure from 2D images only (\refFig{Teaser}).
Reasoning about the 3D structure from 2D observations without assuming anything about their relation is challenging as illustrated by Plato's Allegory of the Cave \cite{warmington1956great}: \textit{How can we hope to understand higher dimensions from only ever seeing projections?}
If multiple views (maybe only two \cite{zhou2017unsupervised,godard2017unsupervised}) of the same object are available, multi-view analysis without 3D supervision has been successful. Regrettably, most photo collections do not come in this form but are now and will remain \textit{unstructured}: they show random instances under random pose, uncalibrated lighting in unknown relations, and multiple views of the same objects are not available. 

\myfigurePlace{Teaser}{
\textsc{PlatonicGAN}s allow converting an unstructured collection of 2D images of a rare class (subset shown on top) into a generative 3D model (random samples below).
}{b}

\mycfigure{Overview}{
Overview: We encode a 2D input image using an encoder $E$ into a latent code $\mathbf z$ and feed it to a generator $G$ to produce a 3D volume.
This 3D volume is inserted into a rendering layer $R$ to produce a 2D rendered image which is presented to a discriminator $D$.
The rendering layer is controlled by an image formation model: visual hull (VH), absorption-only (AO) or emission-absorption (EA) and view sampling. 
The discriminator $D$ is trained to distinguish such rendered imagery from an unstructured 2D photo collection, \ie images of the same class of objects, but not necessarily having repeated instances, view or lighting and with no assumptions about their relation (e.g., annotated feature points, view specifications).
}

Our first main contribution (\refSec{OurApproach}) is to use adversarial training of a 3D generator with a discriminator that operates exclusively on widely available unstructured collections of 2D images, which we call \emph{platonic discriminator}.
Here, during training, the generator produces a 3D shape that is projected (rendered) to 2D and presented to the 2D Platonic discriminator.
Making a connection between the 3D generator and the 2D discriminator, our second key contribution, is enabled by a family of \emph{rendering layers} that can account for occlusion and color (\refSec{RenderingLayers}).
These layers do not need any learnable parameters and allow for backpropagation \cite{rumelhart1988learning}.
From these two key blocks we construct a system that learns the 3D shapes of common classes such as chairs and cars, but also exotic classes from unstructured 2D photo collections. \\
We demonstrate 3D reconstruction from a single 2D image as a key application (\refSec{Evaluation}). While recent works focus on using as little explicit supervision \cite{kanazawa2018category, kato2018neural, eslami2018neural, tulsiani2017multi, tulsiani2018multi, gadelha2016prgan} as possible, they all rely on either annotations, 3D templates, known camera poses, specific views or multi-view images during training. Our approach takes it a step further by receiving no such supervision, see \refTbl{Tickboxes}.

\begin{table}[thb]

\setlength{\tabcolsep}{2.5pt}
    \centering
    \caption{Taxonomy of different methods that learn 3D shapes with no explicit 3D supervision. We compare Kanazawa et al. \cite{kanazawa2018category}, Kato et al. \cite{kato2018neural}, Eslami et al. \cite{eslami2018neural}, Tulsiani et al. \cite{tulsiani2017multi}, Tulsiani et al. \cite{tulsiani2018multi}, PrGan \cite{gadelha2016prgan} with our method in terms of degree of supervision. }
    \label{tbl:Tickboxes}
    \begin{tabular}{l cccccccc}
    Supervision at training time &
    \rot{90}{1em}{\cite{kanazawa2018category}}&
    \rot{90}{1em}{\cite{kato2018neural}}&
    \rot{90}{1em}{\cite{eslami2018neural}}&
    \rot{90}{1em}{\cite{tulsiani2017multi}}&
    \rot{90}{1em}{\cite{tulsiani2018multi}}&
    \rot{90}{1em}{\cite{gadelha2016prgan}}&
    \rot{90}{0em}{Ours}
    \\
    \toprule
Annotation-free & \xmark & \cmark & \cmark & \cmark & \cmark & \cmark & \cmark \\
3D template-Free & \xmark & \xmark & \cmark & \cmark & \cmark & \cmark & \cmark \\
Unknown camera pose & \cmark & \xmark & \xmark & \xmark & \cmark & \cmark & \cmark \\
No pre-defined camera poses & \cmark & \cmark & \cmark & \cmark & \xmark & \xmark & \cmark \\
Only single view required & \cmark & \xmark & \xmark & \xmark & \xmark & \cmark & \cmark \\
Color & \cmark & \cmark & \cmark & \cmark & \xmark & \xmark & \cmark \\
    \bottomrule
    \end{tabular}
\end{table}

\mysection{Related Work}{RelatedWork}

Several papers suggest (adversarial) learning using 3D voxel representations \cite{wu20153d,wu2016learning,girdhar2016learning,qi2016volumetric,gadelha2016prgan,wang2017shape,wang2018deep, wu2017marrnet,yang20173d,varley2017shape, kazhdan2013screened} or point cloud input \cite{achlioptas2018learning, fan2017point}.
The general design of such networks is based on an encoder that generates a latent code which is then fed into a generator to produce a 3D representation (\ie a voxel grid).
A 3D discriminator now analyzes samples both from the generator and from the ground truth distribution.
Note that this procedure requires 3D supervision, \ie is limited by the type and size of the 3D data set such as ShapeNet \cite{shapenet2015}.

Girdhar~\etal\cite{girdhar2016learning} work on a joint embedding of 3D voxels and 2D images, but still require 3D voxelizations as input.
Fan~\etal
\cite{fan2016point} produce points from 2D images, but similarly with 3D data as training input.
Gadelhan~\etal\cite{gadelha2016prgan} use 2D visual hull images to train a generative 3D model. 
Cho~\etals\ recursive design takes multiple images as input \cite{choy20163d} while also being trained on 3D data.
Kar~\etal\cite{kar2017learning} propose a simple ``unprojection'' network component to establish a relation between 2D pixels and 3D voxels but without resolving occlusion and again with 3D supervision.

Cashman and Fitzgibbon~\cite{cashman2013shape} and later Carreira~\etal\cite{carreira2016lifting} or Kanazawa~\etal\cite{kanazawa2018category} use correspondence to 3D templates across segmentation- or correspondence-labeled 2D image data sets to reconstruct 3D shapes.
These present stunning results, for example on animals, but at the opposite end of a spectrum of manual human supervision, in which our approach receives no such supervision.

Closer to our approach is Rezende~\etal\cite{rezende2016unsupervised} that also learn 3D representations from single images.
However, they make use of a partially differentiable renderer \cite{Loper2014OpenDR} that is limited to surface orientation and shading, while our formulation can resolve both occlusion from the camera and appearance.
Also, their representation of the 3D volume is a latent one, that is, it has no immediate physical interpretation that is required in practice, \eg for measurements, to run simulations such as renderings or 3D printing.
This choice of having a deep representation of the 3D world is shared by Eslami~\etal\cite{eslami2018neural}.
Tulsiani~\etal\cite{tulsiani2017multi} reconstruct 3D shape supervised by multiple 2D images of the same object with known view transformations at learning time. Tulsiani~\etal\cite{tulsiani2018multi} take it a step further and require no knowledge about the camera pose, but still require multiple images of the same object at training time. They have investigated modelling image formation as sums of voxel occupancies to predict termination depth. 
We use a GAN to train on photo collections which typically only show one view of each instance.
Closest to our work is Gadelha~\etal\cite{gadelha2016prgan} which operates on an unstructured set of visual hull images but receives three sources of supervision:
view information gets explicitly encoded as a dimension in the latent vector; views come from a manually-chosen 1D subspace (circle); and there are only 8 discrete views.
We take the image formation a step further to support absorption-only and emission-absorption image formation, allowing to learn from real photos and do so on unstructured collections from-the-wild where no view supervision is available.

While early suggestions how to extend differentiable renderers to polygonal meshes exist, they are limited to deformation of a pre-defined template \cite{kato2018neural}.
We work with voxels, which can express arbitrary topology, \eg we can generate chairs with drastically different layout, which are not a mere deformation of a base shape.

Similarly, inter-view constraints can be used to learn depth maps \cite{zhou2017unsupervised,godard2017unsupervised} using reprojection constraints: If the depth label is correct, reprojecting one image into the other view has to produce the other image.
Our method does not learn a single depth map but a full voxel grid and allows principled handling of occlusions.

A generalization from visual hull maps to full 3D scenes is discussed by  Yan~\etal\cite{yan2016perspective}. Instead of a 3D loss, they employ a simple projection along major axis allowing to use a 2D loss.
However, multiple 2D images of the same object are required.
In practice this is achieved by rendering the 3D shape into 2D images from multiple views.
This makes two assumptions: We have multiple images in a \textit{known} relation and available reference appearance (\ie light, materials).
Our approach relaxes those two requirements: we use a discriminator that can work on arbitrary projections and arbitrary natural input images, without known reference.

\mysection{3D Shape From 2D Photo Collections}{OurApproach}
We now introduce \name\ (\refFig{Overview}). The rendering layers used here will be introduced in \refSec{RenderingLayers}.

\paragraph{Common GAN}
Our method is a classic (generative) adversarial design \cite{goodfellow2014generative} with two main differences: The discriminator $ D$ operates in 2D while the 3D generator $G$ produces 3D output. The two are linked by a fixed-function projection operator, \ie non-learnable (see \refSec{RenderingLayers}).

Let us recall the classic adversarial learning of 3D shapes \cite{wu2016learning}, which is a min-max game
\begin{align}
\min_\Theta
\max_\Psi \
c_\mathrm{Dis}(\Psi) +
c_\mathrm{Gen'}(\Theta)
\end{align}
between the discriminator and the generator cost, respectively $c_\mathrm{Dis}$ and $c_\mathrm{Gen'}$.

The discriminator cost is
\begin{align}
c_\mathrm{Dis}(\Psi) = & 
\mathbb{E}_{p_\mathrm{Data}(\mathbf x)}
[\log(D_\Psi(\mathbf x))]
\end{align}
where $D_\Psi$ is the discriminator with learned parameters $\Psi$ which is presented with samples $\mathbf x$ from the distribution of real 3D shapes $\mathbf x\sim p_\mathrm{Data}$.
Here $\mathbb{E}_p$ denotes the expected value of the distribution $p$.

The generator cost is
\begin{align}
c_\mathrm{Gen'}(\Theta) &= 
\mathbb{E}_{p_\mathrm{Gen}(\mathbf z)}
[\log(1-D_{\Psi}(G_\Theta(\mathbf z))]
\end{align}
where $G_\Theta$ is the generator with parameters $\Theta$ that maps the latent code $\mathbf z\sim p_\mathrm{Gen}$ to the data domain.

\paragraph{\name} The discriminator cost is calculated identical to the common GAN with the only difference that the input samples are rendered 2D images with generation cost
\begin{align}
c_\mathrm{Gen}(\Theta)=
\mathbb{E}_{p_\mathrm{Gen}(\mathbf z)}
\mathbb{E}_{p_\mathrm{View}(\mathbf \omega)}
[\log(1-D_{\Psi}(R(\omega, G_\Theta(\mathbf z)))], \label{eq:PlatonicGAN}
\end{align}
where $ R $ projects the generator result $G_\Theta(\mathbf z)$ from 3D to 2D along the sampled view direction $\omega$. See \refSec{Optimization} for  details.

While many parameterizations for views are possible, we choose an orthographic camera with fixed upright orientation that points at the origin from an Euclidean position $\omega\in\mathbb{S}^2$ on the unit sphere. 
$\mathbb{E}_{p_\mathrm{View}(\mathbf \omega)}$ is the expected value across the distributions $\omega\sim p_\mathrm{View}$ of views.

\paragraph{\name 3D Reconstruction}
Two components in addition to our Platonic concept are required to allow for 3D reconstruction, resulting in
\begin{align}
\min_\Psi
\max_{\Theta, \Phi} \
c_\mathrm{Disc}(\Psi) +
c_\mathrm{Gen}(\Theta,\Phi) +
\lambda c_\mathrm{Rec}(\Theta, \Phi),
\end{align}
where $c_\mathrm{Gen}$ includes an encoding step and $c_\mathrm{Rec}$ encourages the encoded generated-and-projected result to be similar to the encoder input where $ \lambda = 100 $. We detail both of these steps in the following paragraphs:

\paragraph{Generator}
The generator $G_\Theta$ does not directly work on a latent code $\mathbf z$, but allows for an encoder $E_\Phi$ with parameters $\Phi$ that encodes a 2D input image $\mathbf I$ to a latent code $\mathbf z = E_\Phi(\mathbf I)$. The cost becomes, 
\begin{align}  
c_\mathrm{Gen}&(\Theta,\Phi) = \nonumber \\ 
& \mathbb{E}_{p_\mathrm{Dat}(\mathbf I)}
\mathbb{E}_{p_\mathrm{View}(\mathbf \omega)}
[\log(1-D_{\Psi}(R(\omega, G_\Theta( E_\Phi(\mathbf I) )))] \label{eq:fullPlatonicGAN}.
\end{align}

\paragraph{Reconstruction}
We encourage the encoder $E_\Phi$ and generator $G_\Theta$ to reproduce the input in the $\mathcal L_2$ sense: by convention the input view is $\omega_0 = (0,0)$, 
\begin{align}
c_\mathrm{Rec}(\Theta, \Phi) = 
\|\mathbf y - R(\omega_0, G_\Theta(E_\Phi(\mathbf I)))\|^2_2
\end{align}
where $ \mathbf y $ represents the ground truth image.
While this step is not required for generation it is mandatory for reconstruction. Furthermore, it adds stability to the optimization as it is easy to find an initial solution that matches this 2D cost before refining the 3D structure.

\mysubsection{Optimization}{Optimization}
Two key properties are essential to successfully optimize our \name: First, maximizing the expected value across the distribution of views $p_\mathrm{View}$ and second, back-propagation through the projection operator $R$. We extend the classic GAN optimization procedure in \refAlg{Ours}. \\

\begin{algorithm}[htb]
    \caption{\name Reconstruction Update Step}
    \label{alg:Ours}
    \begin{algorithmic}[1]
        \State $I_\mathrm{Dat}
        \gets 
        \Call{sampleImage}{p_\mathrm{Dat}}$
        
        \State $\omega 
        \gets 
        \Call{sampleView}{p_\mathrm{View}}$
        
        \State $z 
        \gets 
        \Call{E}{I_\mathrm{Dat}}$
        
        \State $ v
        \gets
        \Call{G}{z}$

        \State $I_\mathrm{View}
        \gets 
        \Call{R}{\omega, v}$
        
        \State $I_\mathrm{Front}
        \gets 
        \Call{R}{\omega_0, v}$
    
        \State $c_\mathrm{Dis}
        \gets 
        \log  D(I_\mathrm{Dat}) +  \log  (1 - D(I_\mathrm{View}))$
        
        \State $c_\mathrm{Gen}
        \gets 
        \log (1 - D(I_\mathrm{View})) $
        
        \State $c_\mathrm{Rec}
        \gets 
        \Call{L2}{I_\mathrm{Dat} - I_\mathrm{Front}}$
        
        \State $\Psi
        \gets
        \Call{maxmize}{c_\mathrm{Dis}}$
        
        \State $\Theta, \Phi
        \gets
        \Call{minimize}{c_\mathrm{Gen} + \lambda c_\mathrm{Rec}}$ 
    
    \end{algorithmic}
\end{algorithm}

\paragraph{Projection}
We focus on the case of a 3D generator on a regular voxel grid $ \mathbf v^{n_\mathrm c\times n_\mathrm p^3} $ and a 2D discriminator on a regular image $ \mathbf I^{n_\mathrm c \times n_\mathrm p^2} $ where $n_\mathrm c$ denotes the number of channels and $n_\mathrm p = 64$ corresponds to the resolution.
In section~\ref{sec:RenderingLayers},  we discuss three different projection operators.
We use $R(\mathbf \omega, \mathbf v)$ to map a 3D voxel grid $ \mathbf v $ under a view direction $ \omega \in \mathbb{S}^2 $ to a 2D image $ \mathbf I $. 

We further define $
R(\mathbf \omega,\mathbf v)
:=
\rho(\mathsf T(\mathbf \omega) \mathbf v)
$ with rotation matrix $\mathsf T(\mathbf\omega)$ according to the view direction $\omega$  and an image formation function $\rho(\mathbf v)$ that is view-independent.
The same transformation is shared by all implementations of the rendering layer, so we will only discuss the key differences of $\rho$ in the following. 
Note that a rotation and a linear resampling is back-propagatable and typically provided in a deep learning framework, \eg as \texttt{torch.nn.functional.grid\_sample} in PyTorch \cite{paszke2017automatic}.
While we work in orthographic space, $\rho$ could also model a perspective transformation.

\paragraph{View sampling} We assume uniform view sampling.

\mysection{Rendering Layers}{RenderingLayers}

Rendering layers (\refFig{RenderingLayer}) map 3D information to 2D images so they can be presented to a discriminator.
We first assume the 3D volume to be rotated (\refFig{RenderingLayer}, a) into camera space from view direction $\omega$ (\refFig{RenderingLayer}, b), such that the pixel value $p$ is to be computed from all voxel values $\mathbf v_i$ and only those (\refFig{RenderingLayer}, c).
The rendering layer maps a sequence of $n_\mathrm z$ voxels to a pixel value $
\rho (\mathbf v)
\in
\mathbb R^{n_\mathrm c \times n_\mathrm p^3}
\rightarrow
\mathbb R^{n_\mathrm c \times n_\mathrm p^2}
$.
Composing the full image $\mathbf I$ just amounts to executing $\rho$ for every pixel $p$ resp.\ all voxels $\mathbf v=v_1,\ldots,v_{n_\mathrm z}$ at that pixel.

\myfigure{RenderingLayer}{Rendering layers (Please see text).}

Note, that the rendering layer does not have any learnable parameters.
We will now discuss several variants of $\rho$, implementing different forms of volume rendering \cite{drebin1988volume}. \refFig{ImageFormations} shows the image formation models we currently support.

\myfigure{ImageFormations}{Different image formation models visual hull (VH), absorption-only (AO) and emission-absorption (EA).}

\paragraph{Visual hull (VH)}
Visual hull \cite{laurentini1994visual}  is the simplest variant (\refFig{ImageFormations}).
It converts scalar density voxels into binary opacity images. 
A voxel value of 0 means empty space and a value of 1 means fully occupied, \ie $v_i \in [0,1]$. Output is a binary value indicating if any voxel blocked the ray.
It is approximated as
\begin{align}
\rho_\text{VH}(\mathbf v)
=
1-\exp(\sum_i -v_i).
\end{align}
Note that the sum operator can both be back-propagated and is efficiently computable on a GPU using a parallel scan.
We can apply this to learn 3D structure from binary 2D data such as  segmented 2D images.

\paragraph{Absorption-only (AO)}
The absorption-only  model is the gradual variant of visual hull.
This allows for ``softer'' attenuation of rays.
It is designed as:
\begin{align}
\rho_\text{AO}(\mathbf v)
=
1-\prod_i (1-v_i).
\end{align}
If $v_i$ are fractional the result is similar to an x-ray, \ie $v_i \in [0,1]$. 
This image formation allows learning from x-rays or other transparent 2D images.
Typically, these are single-channel images, but a colored variant (\eg x-ray at different wavelength or RGB images of colored transparent objects) could technically be done.

\paragraph{Emission-absorption (EA)}
Emission-absorption  allows the voxels not only to absorb light coming towards the observer but also to emit new light at any position.
This interplay of emission and absorption can model occlusion, which we will see is useful to make 3D sense of a 3D world.
\refFig{RenderingLayer} uses emission-absorption with high absorption, effectively realizing an opaque surface with visibility.

A typical choice is to have the absorption $v_\mathrm a$ monochromatic and the emission $v_\mathrm e$ chromatic.

The complete emission-absorption equation is
\begin{align}
\rho_\text{EA}(\mathbf v)
=
\sum_{i=1}^{n_\mathrm z}
\underbrace{
(1-
\prod_{j=1}^{i}
(1-v_{\mathrm a,j}))
}_\text{Transmission $t_i$ to voxel $i$}
v_{\mathrm e,i}
\end{align}
While such equations are typically solved using ray-marching \cite{drebin1988volume}, they can be rewritten to become differentiable in practice:
First, we note that the transmission $t_i$ from voxel $i$ is one minus a product of one minus the density of all voxels before $i$.
Similar to a sum such a cumulative product can be back-propagated and computed efficiently using parallel scans, \eg using \texttt{torch.cumprod}.
A numerical alternative, that performed similar in our experiments, is to work  in the log domain and use \texttt{torch.cumsum}.

\mysection{Evaluation}{Evaluation}
Our evaluation comprises of a quantitative (\refSec{QuantitativeEvaluation}) and a qualitative analysis (\refSec{Qualitative}) that compares different previous techniques and ablations to our work (\refSec{Methods}).

\mysubsection{Data sets}{DataSets}

\paragraph{Synthetic}
We evaluate on two synthetic data sets: (a)~ShapeNet \cite{shapenet2015} and (b)~mammalian~skulls \cite{henzler2018singleimagetomography}.
For our quantitative analysis, we use ShapeNet models as 3D ground truth is required, but strictly only for evaluation, never in our training. 2D images of 3D shapes are rendered for the three image formation models \textsc{VH}, \textsc{AO}, \textsc{EA}. Each shape is rendered from a random view (50 per object), with random natural illumination. ShapeNet only provides 3D density volumes which is not sufficient for \textsc{EA} analysis. To this end, we use volumetric projective texturing to propagate the appearance information from thin 3D surface crust as defined by ShapeNet's textures into the 3D voxelization in order to retrieve \textsc{RGBA} volumes where \textsc{A} corresponds to density.
We use shapes from the classes  
\texttt{airplane},
\texttt{car},
\texttt{chair}, 
\texttt{rifle} and
\texttt{lamp}.
The same train / validation / test split as proposed by \cite{shapenet2015} is adopted.

We also train on a synthetic x-ray data set that consists of 466,200 mammalian skull x-rays~\cite{henzler2018singleimagetomography}. We used the monkey skulls subset of that data set ($\sim$30k x-rays).

\paragraph{Real}
We use two data sets of rare classes: (a)~\texttt{chanterelle} (60 images) and (b)~\texttt{tree} (37 images) (not strictly rare, but difficult to 3D-model).
These images are RGBA, masked, on white background.
Note, that results on these input data has to remain qualitative, as we lack the 3D information to compare to and do not even have a second view of the same object to even perform an image comparison.

\begin{table*}[t!]
\setlength{\tabcolsep}{3.75pt}
\caption{
Performance of different methods with varying degrees of supervision (superv.) \textbf{(rows)} on different metrics \textbf{(columns)} for the class \texttt{airplane}.
Evaluation is performed on all three image formations (IF): visual hull (VH), absorption-only (AO) and emission-absorption (EA).
Note, DSSIM and VGG values are multiplied by $10$ and RMSE by $10^2$.
Lower is better except for IoU.
}

\label{tbl:Main}
\center
\begin{tabular}{l ccc rrrrrrrrrr rrr r}
\multicolumn{1}{c}{Method} & 
IF &
\multicolumn{2}{c}{Superv.}&
\multicolumn{10}{c}{2D Image Re-synthesis}&
\multicolumn{3}{c}{3D Volume}&
\multicolumn{1}{c}{FID}
\\
\cmidrule(lr){1-1}
\cmidrule(lr){2-2}
\cmidrule(lr){3-4}
\cmidrule(lr){5-14}
\cmidrule(lr){15-17}
\cmidrule(lr){18-18}
& & & & 
\multicolumn{2}{c}{VH}&
\multicolumn{2}{c}{AO}&
\multicolumn{2}{c}{EA}&
\multicolumn{2}{c}{VOX}&
\multicolumn{2}{c}{ISO}&
& & & 
\multicolumn{1}{c}{EA}
\\
\cmidrule(lr){5-6}
\cmidrule(lr){7-8}
\cmidrule(lr){9-10}
\cmidrule(lr){11-12}
\cmidrule(lr){13-14}
\cmidrule(lr){18-18}
& &
\scalebox{0.6}{2D} & 
\scalebox{0.6}{3D} & 
\scalebox{0.6}{DSSIM} & \scalebox{0.6}{VGG} & \scalebox{0.6}{DSSIM} & \scalebox{0.6}{VGG} & \scalebox{0.6}{DSSIM} & \scalebox{0.6}{VGG} & \scalebox{0.6}{DSSIM} & \scalebox{0.6}{VGG} & \scalebox{0.6}{DSSIM} & \scalebox{0.6}{VGG} & 
\scalebox{0.6}{RMSE} & \scalebox{0.6}{IoU} & 
\scalebox{0.6}{CD} 
\\
\toprule
PrGAN \cite{gadelha2016prgan} & 
\multirow{5}{*}{\rotatebox{90}{VH}} & 
\cmark& 
\xmark&
1.55 & 6.57 & 1.37 & \textbf{4.85} & 1.41 & \textbf{4.63} & 1.68 & 5.41 & 1.83 & 6.15 & \textbf{7.46} & 0.11 & \textbf{0.22} & 207 \\
Ours & &
\cmark&
\xmark&
\textbf{1.14} & \textbf{5.37} & \textbf{1.16} & 4.93 & \textbf{1.12} & 4.68 & \textbf{1.33} & \textbf{5.22} & \textbf{1.28} & \textbf{5.96} & 9.16 & \textbf{0.20} & 0.55 & \textbf{55} \\
\cmidrule(lr){1-1}
\cmidrule(lr){3-18}
Mult.-View \cite{yan2016perspective} & & 
\cmark&
\xmark&
0.87 & 4.89 & 0.80 & 4.31 & 0.90 & 4.07 & 1.38 & 4.83 & 1.21 & 5.56 & 5.37 & 0.36 & \textbf{0.29} & 155\\
3DGAN \cite{wu2016learning} & &
\cmark&
\xmark&
0.83 & 5.01 & \textbf{0.75} & 4.02 & 0.86 & \textbf{3.83} & 1.30 & 4.73 & 1.17 & 5.82 & \textbf{4.97} & \textbf{0.46} & 0.48 & 111\\
Ours 3D & &
\cmark&
\xmark&
\textbf{0.81} & \textbf{4.82} & 0.77 & \textbf{3.98} & \textbf{0.83} & \textbf{3.83} & \textbf{1.18} & \textbf{4.59} & \textbf{1.09} & \textbf{5.50} & 5.20 & 0.44 & 0.42 & \textbf{98}\\
\midrule
PrGAN \cite{gadelha2016prgan} &
\multirow{5}{*}{\rotatebox{90}{AO}}&
\cmark&
\xmark&
1.41 & 6.40 & 1.27 & 4.80 & 1.27 & 4.52 & 1.53 & 5.32 & 1.63 & 6.00 & 7.11 & 0.09 & \textbf{0.16} & 190\\
Ours & & 
\cmark&
\xmark&
\textbf{0.94} & \textbf{5.35} & \textbf{0.93} & \textbf{4.46} & \textbf{0.91} & \textbf{4.26} & \textbf{1.11} & \textbf{4.96} & \textbf{1.09} & \textbf{5.75} & \textbf{5.70} & \textbf{0.27} & 0.36 & \textbf{90}\\
\cmidrule(lr){1-1}
\cmidrule(lr){3-18}
Mult.-View \cite{yan2016perspective} & & 
\cmark&
\xmark&
0.95 & 4.99 & 0.78 & 4.23 & 0.91 & 4.01 & 1.51 & 4.92 & 1.29 & 5.39 & \textbf{4.89} & 0.34 & \textbf{0.28} & 165\\
3DGAN \cite{wu2016learning} & & 
\cmark&
\xmark&
0.67 & 4.37 & 0.69 & 3.77 & 0.72 & 3.57 & 0.99 & \textbf{4.25} & 0.97 & \textbf{4.92} & 5.08 & \textbf{0.43} & 0.50 & \textbf{58}\\
Ours 3D& & 
\cmark&
\xmark&
\textbf{0.66} & \textbf{4.36} & \textbf{0.66} & \textbf{3.73} & \textbf{0.70} & \textbf{3.52} & \textbf{0.98} & 4.28 & \textbf{0.96} & 4.94 & 5.17 & 0.37 & 0.53 & 64\\
\midrule
PrGAN \cite{gadelha2016prgan} & 
\multirow{5}{*}{\rotatebox{90}{EA}}& 
\cmark&
\xmark&
\textbf{1.31} & \textbf{6.22} & \textbf{1.15} & \textbf{4.77} & \textbf{1.16} & \textbf{5.37} & \textbf{1.36} & \textbf{6.71} & \textbf{1.47} & \textbf{7.07} & \textbf{6.80} & 0.08 & \textbf{0.12} & 196\\
Ours & & 
\cmark&
\xmark&
2.18 & 6.53 & 1.99 & 5.38 & 1.89 & 6.00 & 2.21 & 7.43 & 2.36 & 7.92 & 14.13 & \textbf{0.13} & 1.24 & \textbf{181}\\
\cmidrule(lr){1-1}
\cmidrule(lr){3-18}
Mult.-View \cite{yan2016perspective} & & 
\cmark&
\xmark&
1.62 & 6.21 & 1.53 & 4.58 & 1.63 & 5.48 & 1.95 & 6.97 & 1.94 & 7.41 & 15.05 & 0.12 & 2.52 & 172\\
3DGAN \cite{wu2016learning} & & 
\cmark&
\xmark&
0.89 & 5.28 & \textbf{0.78} & \textbf{3.93} & 0.98 & 4.79 & 1.29 & 6.76 & 1.30 & 7.09 & \textbf{5.24} & \textbf{0.46} & \textbf{0.47} & 110\\
Ours 3D & & 
\cmark&
\xmark&
\textbf{0.82} & \textbf{4.71} & 0.82 & 3.96 & \textbf{0.97} & \textbf{4.77} & \textbf{1.12} & \textbf{6.12} & \textbf{1.16} & \textbf{6.47} & 7.43 & 0.04 & 1.10 & \textbf{73}\\

\bottomrule
\end{tabular} 
\end{table*}

\mysubsection{Baselines and comparison}{Methods}

\paragraph{2D supervision}
First, we compare the publicly available implementation of PrGAN \cite{gadelha2016prgan} with our Platonic method. PrGAN is trained on an explicitly created data set adhering to their view restrictions (8 views along a single axis). Compared to our method it is only trained on visual hull images, however for evaluation purposes absorption-only and emission-absorption (in form of luminance) images are used as input images at test time.
Note that PrGAN allows for object-space view reconstruction due to view information in the latent space whereas our method performs reconstruction in view-space. Due to the possible ambiguities in the input images (multiple images can belong to the same 3D volume), the optimal transformation into object space is found using a grid search across all rotations.

\paragraph{3D supervision}
The first baseline with 3D supervision is \textsc{multi-view}, that has training-time access to multiple images of the same object \cite{yan2016perspective} in a known spatial relation.
Note, that this is a stronger requirement than for \name that does not require any structure in the adversarial examples: geometry, view, light -- all change, while in this method only the view changes in a prescribed way. 

The second competitor is a classic 3DGAN \cite{wu2016learning} trained with a  Wasserstein  loss~\cite{arjovsky2017wasserstein} and  gradient penalty~\cite{gulrajani2017improved}. 

To compare \name against methods having access to 3D information, we also propose a variant \textsc{Platonic3D} by adding the \name adversarial loss term (for all images and shapes) to the 3DGAN framework. 

\mysubsection{Evaluation Metrics}{EvaluationMetrics}

\paragraph{2D evaluation measures} Since lifting 2D information to 3D can be ambiguous, absolute 3D measures might not be the best suitable measures for evaluation on our task. For instance, a shift in depth of an object under an orthographic camera assumption will result in a higher error for metrics in 3D, but the shift would not have any effect on a rendered image. Thus, we render both the reconstructed and the reference volume from the same 10 random views and compare their images using SSIM / DSSIM \cite{wang2004image} and VGG16 \cite{simonyan2014very} features. For this re-rendering, we further employ four different rendering methods:
the original (\ie $\rho$) image formation (IF), volume rendering (VOL), iso-surface rendering with an iso-value of $.1$ (ISO) and a voxel rendering (VOX), all under random natural illumination.

\paragraph{3D evaluation measures}
We report root-mean-squared-error (RMSE), intersection-over-union (IoU) and chamfer distance (CD). For the chamfer distance we compute a weighted directional distance:
\[
d_\mathrm{CD}(T, O) = 
\frac{1}{N}
\sum_{p_{i} \in T}
\min_{p_{j} \in O}
w_j
\|
\mathbf p_i - \mathbf p_j
\|_2^2, 
\]
where $T$ and $O$ correspond to output and target volumes respectively, and $w_j$ denotes the density value of the voxel at location $\mathbf p_j$. The weighting makes intuitive sense as our results have scalar values rather than binary values, \ie higher densities get penalized more, and $N$ is the total number of voxels in the volume. We give preference to such a weighting opposed to finding a threshold value for binarization. 

\mysubsection{Quantitative evaluation}{QuantitativeEvaluation}

\refTbl{Main} summarizes our main results for the \texttt{airplane} class.
Concerning the image formation models, we see that the overall values are best for AO, which is expected: VH asks for scalar density but has only a binary image; AO provides internal structures but only needs to produce scalar density; EA is hardest, as it needs to resolve both density and color.
Nonetheless the differences between us and competitors are similar across the image formation models.

\paragraph{2D supervision}
We see that overall, our 2D supervised method outperforms PrGAN for VH and AO. Even though PrGAN was not trained on EA it wins for all metrics against our 2D supervised method. However, it even outperforms the 3D supervised methods 3DGAN and \textsc{multi-view} which demonstrates the complexity of the task itself. However, PrGAN for EA only produces density volumes unlike all other methods that produce RGBA volumes. Comparing our 2D supervised method against the 3D supervised methods we see that overall our method produces competitive results. Regarding \textsc{multi-view} we sometimes even perform better. 

\paragraph{3D supervision}
Comparing our \textsc{Platonic3D} variant to the 3D baselines we observe our method to mostly outperform them for 2D metrics. Not surprisingly our method performs worse for 3D metrics as our approach only operates in 2D.

\mycfigure{ReconstructionResult}{Visual results for 3D reconstruction of three classes (\texttt{airplane}, \texttt{chair}, \texttt{rifle}) from multiple views.}

In \refTbl{Classes} we look into the performance across different classes. 
\texttt{rifle} performs best: the approach learns quickly from 2D that a gun has an outer 3D shape that is a revolute structure.
\texttt{chair} performs worst, likely due to its high intra-class variation.

\begin{table}[htbp]
\setlength{\tabcolsep}{1.7pt}
\centering
\caption{
Reconstruction performance of our method for different image formation models \textbf{(columns)} on different classes \textbf{(rows)}. The error metric is SSIM (higher is better).
}
\label{tbl:Classes}
\begin{tabular}{l rrr rrr rrr}
\multirowcell{2}{Class}
&\multicolumn{3}{c}{VH}
&\multicolumn{3}{c}{AO}
&\multicolumn{3}{c}{EA}
\\
\cmidrule(lr){2-4}
\cmidrule(lr){5-7}
\cmidrule(lr){8-10}
&\textsc{\small VOL}&\textsc{\small ISO}&\textsc{\small VOX}
&\textsc{\small VOL}&\textsc{\small ISO}&\textsc{\small VOX}
&\textsc{\small VOL}&\textsc{\small ISO}&\textsc{\small VOX}\\
\toprule
\texttt{plane}
&0.93
&0.92
&0.93
&0.94
&0.93
&0.93
&0.85
&0.76
&0.77
\\
\texttt{rifle}
&0.95
&0.94
&0.95
&0.95
&0.94
&0.95
&0.90
&0.78
&0.80
\\
\texttt{chair}
&0.86
&0.85
&0.85
&0.86
&0.85
&0.86
&0.80
&0.61
&0.63
\\
\texttt{car}
&.841
&.846
&.851
&.844
&.846
&.850
&.800
&.731
&.743
\\
\texttt{lamp}
&.920
&.915
&.920
&.926
&.914
&.920
&.883
&.790
&.803
\\
\bottomrule
\end{tabular}
\end{table}

In \refTbl{MeshCount} we compare the mean VGG error of 
a vanilla 3D GAN trained only on 3D shapes, 
a Platonic approach accessing only 2D images, and 
\textsc{Platonic3D} that has access to both.
We keep the number of 2D images fixed, and increase the number of 3D shapes available; the horizontal axis in \refTbl{MeshCount}.
Without making use of the 3D supervision, the error of \name remains constant, independent of the number of 3D models.
Like this, we see that a \name (red line) can beat both other approaches in a condition where little 3D data is available (left).
When more 3D data is available, \name (green line) wins over a pure 3D GAN (blue line).
We conclude that adding 2D image information to a 3D corpus helps, and when the corpus is small enough even performs better than 3D-only supervised methods.

\mycfigure{MethodComparison}{Comparison of 3D reconstruction results using the class \texttt{plane} between different forms of supervision \textbf{(columns)} for two different input views \textbf{(rows)}.
\name, in the second column, can reconstruct a plausible plane, but with errors such as a wrong number of engines.
The 3D GAN in the third column fixes this error, but at the expense of slight mode collapse where instances look similar and slightly ``fat''.
Combining a 3D GAN with adversarial rendering as in the fourth row, is closest to the reference in the fifth row.
}

\begin{table}
\caption{Effect of number of 3D shapes and 2D images on learning different methods in terms of mean DSSIM error. Lower is better.}%
\vspace{0.2cm}
\begin{minipage}[b]{0.45\linewidth}%
\centering
\includegraphics*[clip, width = 3.2cm]{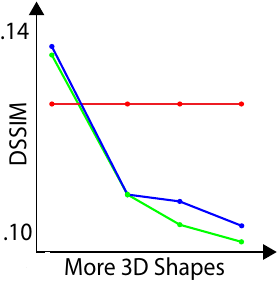}%
\vspace{-1.7cm}%
\end{minipage}%
\begin{minipage}[b]{0.45\linewidth}%
\setlength{\tabcolsep}{1.7pt}
\label{tbl:MeshCount}
\center
\begin{tabular}{l rrrr}
\small  2D images & \small 70k & \small 70k & \small 70k & \small 70k \\
3D shapes & \small 5  & \small 50  & \small 250 & \small 1.5k \\
\midrule
\small  2D-3D ratio & \small 14k & \small 1.4k & \small 280 & \small 47 \\
\toprule
\textcolor{blue}{\textbullet} \small  3D & \small .135 & \small \textbf{.108} & \small .106 & \small .101 \\
\textcolor{red}{\textbullet} \small  Ours & \small \textbf{.125} & \small .125 & \small .125 & \small .125 \\
\textcolor{green}{\textbullet} \small Ours 3D & \small .134 & \small \textbf{.108} & \small \textbf{.102} & \small \textbf{.099} \\
\bottomrule
\end{tabular}%
\end{minipage}%
\end{table}

\mysubsection{Qualitative}{Qualitative}

\paragraph{Synthetic}
\refFig{ReconstructionResult} shows typical results for the reconstruction task. We see that our reconstruction can produce \texttt{airplane}, \texttt{chair} and \texttt{rifle} 3D models representative of the input 2D image.
Most importantly, these 3D models look plausible for multiple views, not only from the input one. The results on the chair category also show that the model captures the relevant variation, ranging from straight chairs over club chairs to armchairs.
For \texttt{gun}, the results turn out almost perfect, in agreement with the numbers reported before.
In summary, our quality is comparable to GANs with 3D supervision.

\paragraph{2D vs.\ 3D vs.\ 2D+3D}
Qualitative comparison of 2D-only, 3D-only and mixed 2D-3D training can be seen in \refFig{MethodComparison}.

\paragraph{Synthetic rare}
We explored reconstructing skulls from x-ray
(\ie the AO IF model) images \cite{henzler2018singleimagetomography} in \refFig{Xray}. We find the method to recover both external and internal structures.

\myfigure{Trees}{3D Reconstruction of different trees using the emission-absorption image formation model, seen from different views \textbf{(columns)}. The small images were used as input. We see that \name has understood the 3D structure, including a distinctly colored stem, fractal geometry and structured leave textures.
}

\paragraph{Real rare}
Results for rare classes are seen in \refFig{Teaser} and
Fig. \refFig{Trees}. We see that our method produces plausible details
from multiple views while respecting the input image, even
in this difficult case. No metric can be applied to these data
as no 3D volume is available to compare in 3D or re-project.

\mysection{Discussion}{Discussion}

\paragraph{Why not having a multi-view discriminator?}
It is tempting to suggest a discriminator that does not only look at a single image, but at multiple views at the same time to judge if the generator result is plausible holistically.
But while we can generate ``fake'' images from multiple views $p_\mathrm{Data}$, the set of ``real'' natural images does not come in such a form. 
As a key advantage, our method only expects unstructured data: online repositories hold  images with unknown camera, 3D geometry or illumination.

\paragraph{Failure cases}{are depicted in \refFig{FailureCases}. Our method struggles to reconstruct the correct pose as lifting 2D images to 3D shapes is ambiguous for view-space reconstruction. }

\myfigure{FailureCases}{Failure cases of a chair \textbf{(top)} and an airplane \textbf{(bottom)}. The encoder is unable to estimate the correct camera-pose due to view-ambiguities in the input image and symmetries in the shapes. The generator then tries to satisfy multiple different camera-poses.}

\myfigure{Xray}{PlatonicGANs trained on 2D x-rays (\ie AO IF) of mammalian skulls \textbf{(a)}.
The resulting 3D volumes can be rendered from novel views using x-ray \textbf{(b)} and under novel views in different appearance, here, using image-based lighting \textbf{(c)}.}

\paragraph{Supplemental}
More analysis, videos, training data and network definitions are available at \url{https://geometry.cs.ucl.ac.uk/projects/2019/platonicgan/}.

\mysection{Conclusion}{Conclusion}
In this paper, we have presented \name, a new approach to learning 3D shapes from unstructured collections of 2D images.
The key to our ``escape plan'' is to train a 3D generator outside the cave that will fool a discriminator seeing projections inside the cave.

We have shown a family of rendering operators that can be GPU-efficiently back-propagated and account for occlusion and color.
These support a range of input modalities, ranging from binary masks, over opacity maps to RGB images with transparency.
Our 3D reconstruction application is build on top of this idea to capture varied and detailed 3D shapes, including color, from 2D images.
Training is exclusively performed on 2D images, enabling 2D photo collections to contribute to generating 3D shapes.

Future work could include shading that is related to gradients of density \cite{drebin1988volume} into classic volume rendering.
Furthermore, any sort of differentiable rendering operator $\rho$ can be added.
Devising such operators is a key future challenge.
Other adversarial applications such as 2D supervised completion of 3D shapes seems worth exploring.
Enabling object-space as opposed to view-space reconstruction would help to prevent failure cases as shown in \refFig{FailureCases}.

While we combine 2D observations with 3D interpretations, similar relations might exist in higher dimensions, between 3D observations and 4D (3D shapes in motion) but also in lower dimensions, such as for 1D row scanner in robotics or 2D slices of 3D data such as in tomography.

\paragraph{Acknowledgements}
This work was supported by the ERC Starting Grant SmartGeometry, a GPU donation by NVIDIA Corporation and a Google AR/VR Research Award.

\clearpage

\small
\bibliographystyle{ieee_fullname}
\bibliography{egbib}

\end{document}